\documentclass[journal,twoside,web]{ieeecolor}
\newcommand{\cD}{\mathcal D}

\newcommand{\cJ}{\mathcal J}

\newcommand{\cL}{\mathcal L}

\newcommand{\cN}{\mathcal N}
\newcommand{\cO}{\mathcal O}

\newcommand{\cX}{\mathcal X}

\newcommand{\bbE}{\mathbb{E}}

\newcommand{\bbP}{\mathbb{P}}

\newcommand{\bbR}{\mathbb{R}}

\newcommand{\vc}{\bm{c}}

\newcommand{\vg}{\bm{g}}

\newcommand{\vl}{\bm{l}}

\newcommand{\vu}{\textbf{u}}

\newcommand{\vx}{\textbf{x}}

\newcommand{\bmat}{\begin{bmatrix}}
\newcommand{\emat}{\end{bmatrix}}

\newcommand{\bsmat}{\begin{bsmallmatrix}}
\newcommand{\esmat}{\end{bsmallmatrix}}

\newcommand{\revised}[1]{\textcolor{black}{#1}}
\newcommand{\irevised}[1]{\textcolor{black}{#1}}

\providecommand{\norm}[1]{\left\lVert#1\right\rVert}

\usepackage{bm}
\usepackage{lcsys}
\usepackage{commath}
\usepackage{comment}
\usepackage{algorithm}
\usepackage[noend]{algpseudocode}
\usepackage{amsmath,amssymb,amsfonts,mathrsfs}

\usepackage{subfigure}
\usepackage{dcolumn}
\usepackage{caption}
\usepackage{makecell}
\usepackage{multirow}
\usepackage{graphicx}
\usepackage{subcaption}
\usepackage{tabularray}
\usepackage{stackengine}
\usepackage{url}
\usepackage{cite}
\usepackage{bigfoot}
\usepackage{textcomp}
\let\labelindent\relax
\usepackage{enumitem}
\usepackage{anyfontsize}
\usepackage[euler]{textgreek}
\usepackage[hidelinks]{hyperref}
\DeclareNewFootnote[para]{default}

\newtheorem{lemma}{Lemma}
\newtheorem{remark}{Remark}
\newtheorem{theorem}{Theorem}

\newtheorem{definition}{Definition}
\newtheorem{assumption}{Assumption}

\def\BibTeX{{\rm B\kern-.05em{\sc i\kern-.025em b}\kern-.08em
    T\kern-.1667em\lower.7ex\hbox{E}\kern-.125emX}}

\begin{document}
\title{\fontsize{22.7}{24}\selectfont Chance-constrained Linear Quadratic Gaussian Games for Multi-robot Interaction under Uncertainty}
\author{Kai Ren, Giulio Salizzoni, Mustafa Emre Gürsoy, and Maryam Kamgarpour	\thanks{Ren, Salizzoni and Kamgarpour are with the SYCAMORE Lab, École Polytechnique Fédérale de Lausanne (EPFL), Switzerland (e-mail: {\tt\small kai.ren@epfl.ch; maryam.kamgarpour@epfl.ch}).}
\thanks{
Ren and Salizzoni's research is supported by Swiss National Foundation Grant $\#200020\_207984 \slash  1$.}
}

\maketitle 
\thispagestyle{empty} 
\begin{abstract}
    We address safe multi-robot interaction under uncertainty. In particular, we formulate a chance-constrained linear quadratic Gaussian game with coupling constraints and system uncertainties. We find a tractable reformulation of the game and propose a dual ascent algorithm. We prove that the algorithm converges to a feedback generalized Nash equilibrium of the reformulated game, ensuring the satisfaction of the chance constraints. We test our method in driving simulations and real-world robot experiments. Our method ensures safety under uncertainty and generates less conservative trajectories than single-agent model predictive control. 
\end{abstract} \vspace{-1mm}
 \vspace{-2mm}

\section{Introduction} \label{sec:introduction}

\IEEEPARstart{I}{n} real-world multi-robot interactive applications, such as warehouse automation, drone swarms and autonomous vehicle fleets, the interaction dynamics play a critical role.  For example, if one robot adopts an aggressive driving style, nearby robots need to yield to ensure safety. Incorporating these mutual influences is essential for achieving safe and efficient multi-robot interactions. To this end, our work addresses safety for multi-agent interaction under uncertainty.

A dynamic game framework \cite{Basar1998} models the sequential interactions among multiple agents \irevised{who jointly control a dynamical system over a time horizon. The decision of one agent at one time step influences the costs of the other agents.} Under the constrained dynamic game setting, generalized Nash equilibrium (GNE) \cite{Facchinei2007} captures the set of equilibria that satisfy the constraints coupling agents' decisions. This framework has gained recent attention in autonomous driving and robotics \cite{cleach2021algames, Zhu2023, Zhong2023}.

Open-loop GNE frameworks compute a sequence of control inputs at the beginning of the game. These frameworks compute the GNE by local first-order algorithms \cite{Di2020} or second-order methods, such as sequential quadratic programming \cite{Zhu2023} and Newton's method \cite{cleach2021algames}. Moreover, special structures of the dynamic games were explored to facilitate convergence and computation. For example, \cite{kavuncu2021potential, bhatt2023efficient} considered the potential dynamic game, where an open-loop GNE can be solved from a single optimal control problem. \revised{Under stochastic uncertainty and chance constraints, \cite{Li2023, yadollahi2025} showed convergence to a subset of open-loop GNE, namely variational equilibria.} 
However, open-loop GNE frameworks lack feedback mechanisms in interactive environments and require receding-horizon implementations to adapt to real-time changes.
 
 A feedback Nash equilibrium can be computed in an unconstrained linear-quadratic (LQ) game via dynamic programming \cite{Basar1998}. For constrained LQ games, the feedback GNE was derived in \cite{Reddy2017, Laine2023}. However, these works assumed a deterministic setting and did not account for stochastic uncertainty, which is crucial in real-world robot interactions.
 
To address safety under uncertainty, recent works considered system noise and risk constraints in single-agent linear quadratic regulator (LQR) problems. For a finite-horizon risk-constrained LQR problem \cite{Tsiamis2020CDC}, a bisection method was employed to compute the optimal feedback policy. Its extension to the infinite-horizon setting \cite{Zhao2021} showed that a primal-dual method converges sublinearly to the optimal feedback policy. Our work extends these studies to a multi-agent chance-constrained linear quadratic Gaussian (LQG) game setting.
 
 Stochastic uncertainty and chance constraints were considered in a multi-agent dynamic game \cite{Zhong2023}, where a feasible control policy was computed via the augmented Lagrangian method. While the algorithm showed efficacy in autonomous driving simulations, the authors did not provide convergence or safety guarantees. However, such guarantees are essential for safety-critical applications.   

Our work aims to formulate a specific class of LQG games motivated by multi-robot interactions and develop conditions for safety and convergence. Our main contributions are
\begin{itemize}
    \item We formulate a chance-constrained LQG game and derive a \revised{conservative} approximation of the chance constraints. 
    \item We prove that, under certain assumptions, a dual ascent algorithm converges sublinearly to the feedback generalized Nash equilibrium, which guarantees the satisfaction of the chance constraint.
    
    \item We test our method in simulations and real-world robot experiments, demonstrating that our algorithm ensures safety in multi-agent interactions under uncertainty. Our method supports real-time online computation and generates less conservative trajectories compared to single-agent model predictive control.
\end{itemize}

\textit{Notations:} A Gaussian distribution with mean $\mu$ and covariance matrix $\Sigma$ is denoted as $\mathcal{N}(\mu, \Sigma)$. We denote a set of consecutive integers by $[a]=\{1, 2, \dots, a\}$ and $[a]^- = \{0, 1, \dots, a-1\}$.  
The weighted \(L_2\) norm of a vector \( x \) with a weighting matrix \( P \succeq 0\) is $\|x \|_{P} = \sqrt{x^\top P x}$. A conjunction (logical and) is denoted as $\bigwedge$.

\section{Problem Formulation} \label{sec:CCgameProblem}
We consider an $N$-agent dynamic system, which is modeled as a stochastic linear time-varying system with Gaussian noise \vspace{-2mm}
{\small
\begin{equation} \label{eq:systemdynamic}
    x_{t+1} = A_t x_t + \sum_{i=1}^{N} B_{t}^{i} u^i_t + w_t, \;\; w_t \sim \cN (\textbf{0}, W_t), \;\; \forall t \in [T]^-.
\end{equation}} \vspace{-3mm}

We denote the state and control input of agent \( i \) at time \( t \) as \( x^{i}_{t} \revised{\in \mathbb{R}^{n_{x^{i}}}}\) and \( u^{i}_{t} \in \mathbb{R}^{n_u} \) respectively. The shared state vector \( x_t = [x_{t}^{1^\top}, \ldots, x_{t}^{N^\top} ]^\top  \in \mathbb{R}^{n_x} \)\revised{, with $n_{x} = \sum_{i=1}^{N} n_{x^{i}}$,}  contains the states of all agents. For the linear system, $A_t \in \mathbb{R}^{n_x \times n_x}$ and $B_t^i \in \mathbb{R}^{n_x \times n_u}$ are the system's dynamics matrices. We also consider system noise $w_t \in \mathbb{R}^{n_x}$. In this work, we assume $w_t$ is a sample of a Gaussian distribution $w_t \sim \cN(\textbf{0}, W_t)$, with \revised{known} covariance \revised{$W_t \in \bbR^{n_x \times n_x}$ and }$W_t \succ 0$.

Given a known initial state $x_0$ and a horizon $T$, each agent generates a sequence of control inputs $\vu^i = [(u^{i}_{0})^\top, \ldots, (u^{i}_{T-1})^\top]^\top$. Given the system dynamics \eqref{eq:systemdynamic}, the input of all agents would lead to a state trajectory $[x_0, x_1, \dots, x_{T}]$. We denote the expected state trajectory as $\vx = [x_0^\top, \; \bbE(x^{\top}_{1}), \ldots, \bbE(x^{\top}_{T})]^\top$.

To ensure collision avoidance and satisfying constraints such as staying within lanes under system uncertainty, the state of the multi-agent system is confined within a safe set: $\cX := \bigcap_{k=1}^K \cX^k$ with a prescribed probability $\epsilon$, i.e., \vspace{-1mm}
\begin{equation} \label{constraint:jointChance}
\mathbb{P}\left(\textstyle \bigwedge_{t=1}^{T} \displaystyle x_{t} \in \mathcal{X} \right) = \mathbb{P}\left(\textstyle \bigwedge_{t=1}^{T} \bigwedge_{k=1}^{K}\displaystyle x_{t} \in \mathcal{X}^k \right) \geq 1 - \epsilon. 
\end{equation}

With a risk allocation satisfying $\sum_{k=1}^K \sum_{t=1}^T \epsilon_t^k = \epsilon$, we can conservatively approximate $\eqref{constraint:jointChance}$ with the following \cite{Ono2008RA}. \vspace{-1mm}
\begin{equation} \label{constraint:chance}
   \bbP(x_t \in \cX^k) \geq 1-\epsilon_t^k, \;\; \forall k \in [K], \; t \in [T].
\end{equation}

We consider the following two forms of safe set $\cX^k$ for robotic applications.
\begin{enumerate}
    \item Decoupled box constraints, e.g., the vehicles should stay within a lane and conform to speed limits: \vspace{-1mm}
\begin{equation}\label{eq:boxStateConstraint}
    \bbP(x_{\text{min}} \leq x_t \leq x_{\text{max}}) \geq 1 - \epsilon_t^k.
\end{equation}
\item Coupled state constraints for collision avoidance, namely, agents maintain at least a pairwise distance of $R$: \vspace{-1mm}
\begin{equation}
    \mathbb{P}\left(\|x^{i}_{t} - x^{j}_{t}\|^2_C \geq R^2\right) \geq 1 - \epsilon_t^k, \;\; \forall j \neq i. \label{constraint:collisionavoidance}
\end{equation}
\end{enumerate}

Each agent aims to minimize its own cost (e.g., fuel consumption or distance to the goal) while satisfying the chance constraints \eqref{constraint:chance}. As the constraints depend on the other agents' strategies, we consider a discrete-time finite-horizon $N$-player general-sum chance-constrained LQG game. Each player $i$ aims to minimize the following cost function:
\begin{subequations} \label{problem:LQGprimal}
\begin{alignat}{2}
& \underset{\gamma^i }{\text{min}}
& & \;\; \cJ^i(x_0, \gamma) := \mathbb{E} \left[ \sum_{t=0}^{T-1} \left( \norm{x_{t+1}}^2_{Q^{i}_{t+1}} + \|u^i_t\|^2_{R^{i}_{t}} \right)   \right] \label{cost:eachAgent}\\
& \text{s.t.} 
&& \;\; \eqref{eq:systemdynamic}, 
\eqref{constraint:chance}. \label{constraint:allxu}
\end{alignat}
\end{subequations}

The weighing matrices for state and input costs are $Q^{i}_{t} \succeq 0$ and $R^{i}_{t} \succ 0$ respectively.  

We aim to find a feedback policy $\gamma^i:= (\gamma^i_0, \gamma^i_1, \ldots, \gamma^i_{T-1})$ for each player, where $\gamma^i_t: \cX \rightarrow \bbR^{n_u}$ determines the control inputs based on the evolving states: $u_t^i = \gamma^i_t(x_t)$. We denote \( \gamma^{-i} \) the control inputs of all players except player \( i \) and the control policy of all players as $\gamma = (\gamma^i, \gamma^{-i})$.

\begin{definition}
A \revised{feedback} generalized Nash equilibrium (GNE) \cite{Laine2023} is a feasible control policy $(\gamma^{i*}, \gamma^{-i*})$ where no player has a feasible unilateral deviation of the control policy that can reduce their cost, i.e., for all players $i \in [N]$,
\end{definition} \vspace{-4mm}

\begin{equation*}
\begin{aligned}
   &\cJ^i(x_0, (\gamma^{i*}, \gamma^{-i*})) \leq \cJ^i(x_0, (\gamma^i, \gamma^{-i*})), \; \forall \gamma^i ,\\ 
   &\text{s.t. } (\gamma^i, \gamma^{-i*}) \text{ satisfies } \eqref{constraint:allxu}.
\end{aligned}
\label{eq:GNE}
\end{equation*}

Previous works \cite{Reddy2017, Laine2023} on \revised{computing feedback GNE of constrained dynamic games} did not consider the uncertainties. While \cite{Zhong2023} addressed a similar chance-constrained setting, it did not provide a guarantee on the satisfaction of the chance constraints. We aim to develop an algorithm that generates control policies satisfying chance constraints \eqref{constraint:chance}. \vspace{-4mm}

\subsection{Tractable approximation of chance constraints}
As the chance constraint for collision avoidance \eqref{constraint:collisionavoidance} involves a non-convex quadratic constraint function, it is generally intractable \cite{Ono2008RA, Nair2024}. Therefore, we first develop a tractable reformulation of these chance constraints.

\begin{lemma} \label{lemma:affineInner}
    The feedback GNE of the following game ensures the satisfaction of the chance constraints \eqref{constraint:chance}. \vspace{-5mm}

{\normalfont
\begin{subequations} \label{problem:primal} 
\begin{alignat}{2} 
& \underset{\gamma^i }{\text{min}} \quad \cJ^i(x_0, \gamma) \label{cost:reformulated}, \quad \text{s.t. }\;\;  \eqref{eq:systemdynamic}, \\
& \quad \quad\quad\quad\quad\quad\quad\quad \quad\quad\text{and} \;\; \vg (\vx) := \vl^{\top} \vx + \vc  \leq 0. \label{constraint:primalAffine}
\end{alignat}
\end{subequations}}
\end{lemma} \vspace{2mm}

The derivations of $\vl \in \bbR^{T n_x \times M}$ and $\vc \in \bbR^M$, and the proof of Lemma~\ref{lemma:affineInner} is shown in appendix~\ref{appendix:chanceLinear}. The original chance-constrained LQG game \eqref{problem:LQGprimal} has now been reformulated as an LQG game with linear state constraints. \vspace{-2mm}

\section{Solving the Constrained LQG Games} \label{sec:methods}
In this section, we first characterize the feedback generalized Nash equilibrium (GNE) of \eqref{problem:primal} using duality. Then, we present a dual ascent algorithm and prove its convergence to a feedback GNE of \eqref{problem:primal}. \vspace{-4mm}

\subsection{Characterizing the feedback GNE control policy}
Given that player $i$ has cost $\cJ^i$ and constraints \eqref{constraint:primalAffine}, the Lagrangian of player $i$ is \vspace{-2mm}
\begin{equation*}
    \cL^i(\gamma, \bm{\lambda}) := \cJ^i(x_0, \gamma) + {\bm{\lambda}^i}^\top \vg(\vx)
\end{equation*}

In the above, we have denoted the dual variable of player $i$ as ${\bm{\lambda}^i} \in \bbR^M$. Note that the constraints \eqref{constraint:primalAffine} are shared among all agents. Therefore, we consider a class of game where the dual variables are equal over all agents: ${\bm{\lambda}^i} = {\bm{\lambda}^j} = \bm{\lambda}, \; \forall i, j \in [N]$, similarly as in \cite{Zhu2023, cleach2021algames}. From this point onward, we will use $\bm{\lambda}$ to denote the shared dual variables. The primal and dual problems for player $i$ are \vspace{-2mm}
\begin{align*}
    \min_{\gamma^i }  \max_{\bm{\lambda} \geq 0} \; \cL^i(\gamma^i, \gamma^{-i}, \bm{\lambda}), \tag{$P^i$} \label{eq:piprimal}\\
    \max_{\bm{\lambda} \geq 0} \min_{\gamma^i } \; \cL^i(\gamma^i, \gamma^{-i}, \bm{\lambda}), \tag{$D^i$} \label{eq:dual}
\end{align*}
respectively. The dual function of player $i$ is denoted as $\cD^i(\bm{\lambda}; \gamma^{-i}) := \min_{\gamma^i } \cL^i(\gamma^i, \gamma^{-i}, \bm{\lambda}).$ 
We consider the following assumptions for characterizing the primal-dual solutions corresponding to the feedback GNE of \eqref{problem:primal}.
\begin{assumption} \noindent For all players $i\in [N]$, for any $\gamma^{-i}$, the primal and dual problems have a zero duality gap. \label{assumption:strongduality} \end{assumption}

\begin{remark} 
    Note that for a fixed $\gamma^{-i}$, the primal problem becomes a single-agent constrained linear quadratic regulator problem. Under the Slater's condition, \eqref{eq:piprimal} and \eqref{eq:dual} have a zero duality gap \cite{Tsiamis2020CDC}. To ensure strong duality for any $\gamma^{-i}$ as in assumption~\ref{assumption:strongduality}, a sufficient condition is the strong Slater's condition \cite{Altman2000}, which requires that for any $\gamma^{-i}$, there exists a $\gamma^{i}$, such that {\normalfont $\vg(\vx) < 0$}. Deriving relaxed conditions for strong duality under coupling constraints remains future work.
\end{remark}

Let Assumption~\ref{assumption:strongduality} hold. For all players $i \in [N]$, if $\bm{\lambda}^{*}$ is feasible for the dual problem \eqref{eq:dual}, then \revised{if $\gamma^{i*} $ satisfies \vspace{-1mm} 
{
    \begin{align*}
        & \cL^i(\gamma^{i*}, \gamma^{-i*}, \bm{\lambda}^{*}) = \min_{\gamma^i } \cL^i(\gamma^i, \gamma^{-i*}, \bm{\lambda}^{*}), \;\; \\
        & \bm{\lambda^{*}} \geq 0,  \quad \vg (\vx^*) \leq 0, \quad {\bm{\lambda^{*}}}^\top  \vg (\vx^*) = 0, 
    \end{align*}}then player $i$ reaches optimum given the other players' policies are fixed to the feedback GNE $\gamma^{-i*}$ \cite{ruszczynski2006nonlinear}. It follows that the joint policy $\gamma^*$ is a feedback GNE of \eqref{problem:primal} by definition.} Here, $\vx^*$  is the expected state trajectory induced by policy $\gamma^{*}$ with system dynamics \eqref{eq:systemdynamic}. \vspace{-4mm}

\subsection{Dual ascent algorithm} 

Observe that for an arbitrary $\bm{\lambda}$, \revised{an LQG game is induced in which player $i$ solves the following problem: \vspace{-1mm}
\begin{equation} \label{eq:dualGame}
    \underset{\gamma^i }{\text{min}} \;\; \cL^i(\gamma, \bm{\lambda}) = \cJ^i(x_0, \gamma) + {\bm{\lambda}}^\top \vg(\vx), \;\;\; \text{s.t. } \eqref{eq:systemdynamic}, \vspace{-2mm}
\end{equation}}where the Lagrangian $\cL^i$ is a quadratic cost function with additional linear terms in the state. Hence, under suitable conditions, we can uniquely determine the following linear state-feedback Nash equilibrium (NE) \cite{Basar1998}.  \vspace{-1mm}
\begin{equation} \label{eq:feedbackU}
    \gamma^{i*}_{t(\bm{\lambda})}(x_t) = -K^{i*}_{t} x_t - \alpha^{i*}_{t(\bm{\lambda})}, \;\;\; \forall t \in [T]^-,
\end{equation} \vspace{-5mm}

\noindent
where the feedback parameters $(K^{i*}_t, \alpha^{i*}_t)$ are defined by a dynamic programming recursion (see Appendix~\ref{appendix:LQfeedback} for details). 

The formulations above motivate a dual ascent algorithm as follows to compute the feedback GNE of \eqref{problem:primal}. \vspace{-2mm}

\begin{algorithm}[ht]
\begin{minipage}{0.5\textwidth}
\caption{Dual ascent feedback GNE Algorithm}
\label{alg:constrainedLQGG}
\begin{algorithmic}[1]
\State \textbf{Initialize:} initial state $x_0$, initial dual guess $\bm{\lambda}_{(1)}$.

\For{$l = 1, 2, 3, \ldots k$} \vspace{1mm}

\State $\gamma_{\bm{\lambda}_{(l)}} \leftarrow$ Compute the NE \revised{of \eqref{eq:dualGame}} with $\bm{\lambda}_{(l)}$; \vspace{1mm} \label{algStep:solveNE}

\State $\vx_{(l)}\leftarrow$ Integrate \eqref{eq:systemdynamic} with $w_t = 0, \; \forall t \in [T]^-$; \label{algStep:integrateEx}\vspace{1mm}
\State $\bm{\lambda}_{(l+1)} \leftarrow \max \left (0, \; \bm{\lambda}_{(l)} +  \eta \; \vg (\vx_{(l)}) \right)$. \label{algStep:dualUpdate}
\EndFor

\State $\bm{\bar{\lambda}} \leftarrow \frac{1}{k} \sum_{l=1}^k \bm{\lambda_{(l)}}$;

\State $\gamma_{\bm{\bar{\lambda}}} \leftarrow$ Compute the NE \revised{of \eqref{eq:dualGame}} with $\bm{\bar{\lambda}}$.

\State \Return $(\gamma_{\bm{\bar{\lambda}}}, \bm{\bar{\lambda}})$
\end{algorithmic}
\end{minipage}
\end{algorithm} \vspace{-2mm}

\revised{In algorithm~\ref{alg:constrainedLQGG}, line~\ref{algStep:solveNE} computes a linear state-feedback NE policy \eqref{eq:feedbackU} of the LQG game \eqref{eq:dualGame} with $\bm{\lambda} = \bm{\lambda}_{(l)}$. Line~\ref{algStep:integrateEx} integrates the NE policy $\gamma_{\bm{\lambda}_{(l)}}$ to system dynamic \eqref{eq:systemdynamic} with zero uncertainty to obtain the expected state trajectory. Line~\ref{algStep:dualUpdate} updates the dual variables with projected gradient ascent.}

We now show that \revised{$\gamma_{\bar{\bm{\lambda}}}$ computed in Algorithm~\ref{alg:constrainedLQGG} converges} sublinearly to a feedback GNE of \eqref{problem:primal} under assumptions~\ref{assumption:strongduality}. Let us define $\cD^i (\bm{\lambda})= \cD^i (\bm{\lambda}; \gamma^{-i}_{\bm{\lambda}})$, where the other players' policies are fixed to their Nash equilibrium strategies corresponding to a given $\bm{\lambda}$. Note that $\cD^i (\bm{\lambda})$ is the pointwise infimum of a family of affine functions of $\bm{\lambda}$, it is concave \cite[Theorem 6.3.1]{bazaraa2006nonlinear}. To prove convergence, we first characterize the gradient of the $\cD^i (\bm{\lambda})$ as follows: 
\begin{lemma} \label{lemma:dualGradient} 
    For all $\bm{\lambda} \in \bbR^M_{\geq 0}$, $\nabla \cD^i (\bm{\lambda}) =  \vg (\vx_{\bm{\lambda}}) = \Tilde{L}^\top \bm{\lambda} + \Tilde{\vc}$, where $\vx_{\bm{\lambda}}$ is the expected state trajectory induced by the NE policy $(\gamma^{i}_{\bm{\lambda}}, \gamma^{-i}_{\bm{\lambda}})$ corresponding to $\bm{\lambda}$.
\end{lemma}

The \revised{definitions of $\Tilde{L}, \Tilde{\vc}$} and the proof is provided in Appendix~\ref{appendix:proofTheorem}. By lemma~\ref{lemma:dualGradient}, it follows directly that the gradient of $\cD^i (\bm{\lambda})$ is $L$-Lipschitz with $L = \|\Tilde{L}\|_2$, which establishes our main convergence result.

\begin{theorem} \label{theorem:PDconvergence}
Under assumptions~\ref{assumption:strongduality}, with a constant step size $\eta \in (0, 1/L)$, there exists a constant $\xi > 0$ such that \vspace{-2mm}
$$\textstyle \cD^i(\bm{\lambda}^{*}) - \cD^i \left(\frac{1}{k} \sum_{l=1}^k \bm{\lambda_{(l)}} \right) \leq \xi/k.$$
\end{theorem} \vspace{1mm}
 
\begin{proof}
\revised{From line~\ref{algStep:dualUpdate} of Algorithm~\ref{alg:constrainedLQGG}}, it follows that \vspace{-1mm}
 {\small
\begin{align*} 
    & \textstyle \norm{\bm{\lambda}_{(k+1)} - \bm{\lambda}^{*}}_2^2 \leq \norm{\bm{\lambda}_{(k)} - \bm{\lambda}^{*} + \eta \vg(\vx_{(k)})}_2^2 \\
     = & \textstyle \norm{\bm{\lambda}_{(k)} - \bm{\lambda}^{*}}_2^2 + \eta^{2} \|\nabla \cD^i(\bm{\lambda}_{(k)})\|_2^2  + 2 \eta \left(\vg(\vx_{(k)}) \right)^\top (\bm{\lambda}_{(k)} - \bm{\lambda}^{*}) \\ 
     \leq & \textstyle \norm{\bm{\lambda}_{(k)} - \bm{\lambda}^{*}}_2^2 + (2\eta - 2\eta^2 L)(\cD^i(\bm{\lambda}_{(k)}) - \cD^i(\bm{\lambda}^{*})).
\end{align*} }\vspace{-5mm}

The second inequality holds \revised{because $\cD^i(\bm{\lambda})$ is concave and $\nabla\cD^i(\bm{\lambda})$ is $L$-Lipschitz.} The latter gives {\small$\|\nabla \cD^i(\bm{\lambda}_{(k)})\|_2^2 \leq 2L(\cD^i(\bm{\lambda}^{*}) - \cD^i(\bm{\lambda}_{(k)}))$}. Choosing $\eta \in (0, \frac{1}{L})$, we get {\small $\cD^i(\bm{\lambda}^{*}) - \cD^i(\bm{\lambda}_{(k)}) \leq$} {\small $\frac{1}{2\eta - 2\eta^2 L} \left(\norm{\bm{\lambda}_{(k)} - \bm{\lambda}^{*}}_2^2 - \norm{\bm{\lambda}_{(k+1)} - \bm{\lambda}^{*}}_2^2 \right)$.} \vspace{1mm}

Summing up over all iterations $l = 1, 2, \ldots, k$, we get \vspace{-2mm}
{\footnotesize
    \begin{align*}
        & \sum_{l=1}^k \left(\cD^i(\bm{\lambda}^{*}) - \cD^i(\bm{\lambda}_{(l)})\right) \leq  \frac{1}{2 \eta- 2\eta^2 L} \|\bm{\lambda}_{(1)} - \bm{\lambda}^{*}\|_2^2.
    \end{align*}}\vspace{-2mm}

Next, applying the Jensen's inequality, we get \vspace{-1mm}
{\small
\begin{align*}
    \textstyle \cD^i(\bm{\lambda}^{*}) - \cD^i \left(\frac{1}{k} \sum_{l=1}^k \bm{\lambda_{(l)}} \right) & \textstyle \leq \frac{1}{k}\sum_{l=1}^k \left(\cD^i(\bm{\lambda}^{*}) - \cD^i(\bm{\lambda}_{(l)})\right) \\
     & \textstyle \leq \textstyle \frac{1}{2k\eta (1-\eta L)} \|\bm{\lambda}_{(1)} - \bm{\lambda}^{*}\|_2^2.
\end{align*} }

Using a constant step size $\eta = h/L$ with $h \in (0, 1)$, we have $\cD^i(\bm{\lambda}^{*}) - \cD^i \left(\frac{1}{k} \sum_{l=1}^k \bm{\lambda_{(l)}} \right) \leq \frac{L\|\bm{\lambda}_{(1)} - \bm{\lambda}^{*}\|^2_2}{2h(1-h) k}$.
\end{proof} 
 
 Previous work \cite{Zhong2023} used an augmented Lagrangian method to solve the chance-constrained dynamic games, but did not provide a convergence guarantee. While \cite{Zhao2021} showed an $\cO(1/\sqrt{k})$ convergence for a single-agent risk-constrained LQR problem, it required a strong assumption of bounded dual during the iterates. \revised{By proving that the gradient of $\cD^i (\bm{\lambda})$ is Lipschitz in the linear-constrained LQ game (Lemma~\ref{lemma:dualGradient})}, we relax that assumption and achieve an $\cO(1/k)$ convergence rate to the feedback GNE of a multi-player game.

\begin{remark}
    \revised{Based on \eqref{eq:feedbackU} and the proof of Lemma~\ref{lemma:dualGradient} (Appendix~\ref{appendix:proofTheorem}), the feedback parameter stacked over $t\in[T]^-$: $\alpha^* = G \bm{\lambda}$ for some $G \in \bbR^{Tn_u \times M}$, which is continuous in $\bm{\lambda}$. Hence, the convergence of $\bar{\bm{\lambda}}$ implies the convergence of $\gamma_{\bar{\bm{\lambda}}}$ to the feedback GNE $\gamma^*$.} The feedback GNE of \eqref{problem:primal} ensures the satisfaction of \eqref{constraint:primalAffine}, which are conservative approximations of \eqref{constraint:chance} (see Lemma~\ref{lemma:affineInner}). Hence, the policy computed by Algorithm \ref{alg:constrainedLQGG} is asymptotically guaranteed to satisfy the original chance constraints \eqref{constraint:chance}. However, within a finite number of iterations, we cannot guarantee convergence to the feedback GNE or satisfaction of \eqref{constraint:chance}. Developing finite-iteration convergence or bounding the constraint violation gap remains as future work.
\end{remark}

\section{Case Studies} \label{sec:casestudy}\vspace{1mm}

\begin{figure*}[!ht]
    \centering
    \includegraphics[width=0.95\textwidth]{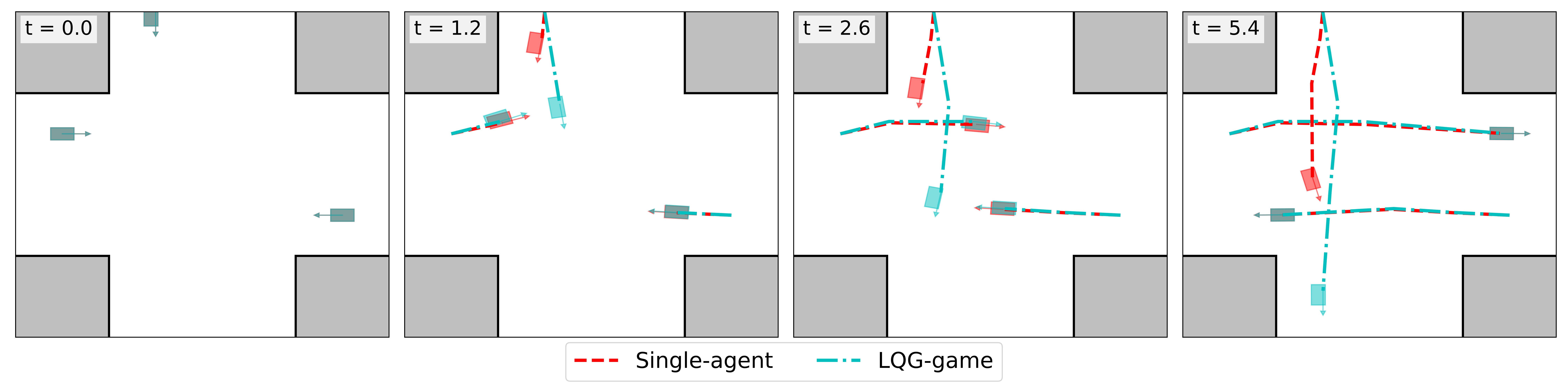}
    \vspace{-2mm}
    \caption{Simulation in a cross-intersection scenario. For the LQG-game planner (Algorithm~\ref{alg:constrainedLQGG}) \revised{and central MPC}, the three-car system reaches its goals efficiently: the vehicle coming from the top crosses the intersection before the vehicles from the left and right while avoiding collisions. In contrast, the single-agent MPC planner \cite{ren2024safe} approaches cautiously: the car from the top waits for the left and right vehicles to pass before crossing, leading to safer but less efficient trajectories.} 
    \label{fig:game} \vspace{-2mm}
\end{figure*} \vspace{-1mm}

In this section, we first test our method in an autonomous driving scenario. We compare our method with a single-agent \revised{and a central} model predictive control (MPC) framework in terms of safety, optimality and computational time. Our goal is to verify whether our game-theoretic method can overcome the conservativeness in the single-agent MPC by accounting for interdependent strategic decisions and shared constraints in multi-agent interactions. Then, we test our method in a multi-robot experiment to examine the real-world performance.

To model the robots' dynamics, we linearize a unicycle model around a nominal trajectory \cite{Keil2020}, resulting in a linear time-varying system~\eqref{eq:systemdynamic}. The state $x_t^i = [p^i_{t, x}, p_{t, y}^i, \theta^i_t, v^i_t]^\top \in \bbR^4$ represents the 2D position, orientation, and linear velocity, while the input $u^i = [a^i_t, \omega^i_t]^\top \in \bbR^2$ consists of the linear acceleration and angular velocity. The cost of each agent \eqref{cost:eachAgent} consists of a reference cost penalizing the terminal state distance to a goal state, and an input cost to promote smooth driving behavior. 

We enforce lane-keeping and speed limit constraints as in \eqref{eq:boxStateConstraint}. The collision avoidance constraint is enforced pairwise between agents as in \eqref{constraint:collisionavoidance}. We consider a risk tolerance $\epsilon=0.05$ with a uniform risk allocation\footnote{One can consider optimizing risk allocation \cite{Ono2008RA,Nair2024} for a less conservative approximation of \eqref{constraint:jointChance}, but it is not a focus of this work.}.  
We consider a 10s horizon with a 0.2s interval, yielding $T=50$ planning timesteps. \vspace{-4mm}

\subsection{Multi-agent autonomous driving simulation} 
We first test our algorithm in an autonomous driving scenario\footnote{The code for the autonomous driving test scenario is available at \href{https://github.com/renkai99/safe-CCLQGGame}{\texttt{https://github.com/renkai99/safe-CCLQGGame}}.}. As shown in Fig.~\ref{fig:game}, at $t=0$, three vehicles arrive at a cross intersection and aim to cross the intersection as soon as possible while avoiding collision with each other.

We compare our algorithm with a single-agent \revised{and a centralized} chance-constrained MPC method (detailed below).

\textbf{(LQG-game) Chance-constrained LQG game:} we employed Algorithm~\ref{alg:constrainedLQGG} to find the state-feedback policy $\gamma_{\bm{\bar{\lambda}}}$.

\revised{\textbf{(Central-MPC)}: we consider the multi-agent system as a single agent. The cost function is the summation of the costs of all agents, i.e., at every planning step we solve for $\vu = [(\vu^{1})^\top, \ldots, (\vu^{N})^\top]^\top$ via $\underset{\vu}{\text{min}}\sum_{i=1}^{N} \cJ^i(x_0, \vu), \; \text{s.t. } \eqref{eq:systemdynamic}, \eqref{constraint:primalAffine}$.} \vspace{1mm}
 
\textbf{(Single-agent) Single-agent MPC \cite{Nair2024,ren2024safe}:} in this setting, only one vehicle (the top car) is controlled, while others are treated as dynamic obstacle vehicles (OVs). The OVs follow the nominal trajectories, with a known state covariance propagation (see Appendix~\ref{appendix:chanceLinear}). We sample 2000 OV trajectories to estimate the Gaussian moments of their uncertain parameters. Based on the predictions, the top car plans its trajectory via chance-constrained MPC \cite[Algorithm 1]{ren2024safe}. 

For all planners, we conducted the simulations 100 times by integrating dynamics \eqref{eq:systemdynamic} with different realizations of the system noise $w_t$. One exemplary set of trajectories from the planners is shown in Fig.~\ref{fig:game}. We compare their performances over the 100 simulations. The results are shown in Table \ref{tab:performance} and summarized below.

\textbf{Optimality:} We evaluate the average cost (the sum of \eqref{cost:eachAgent} for all three cars) and the ``\textbf{Travel T}'' (time taken for all cars to reach the goal) over the 100 simulations. 

The LQG-game \revised{and central-MPC} planners have a lower cost than the single-agent planner. The vehicle from the top driven by the LQG-game \revised{and central-MPC} crosses the intersection before the cars from left and right. For the single-agent planner, the car from the top waits for the left and right cars before entering the intersection.  

\textbf{Collision rate:} We report the collision rate over the 100 trials in the column of \textbf{``Col. rate"} of Table~\ref{tab:performance}. All planners have a collision rate below the threshold of $\epsilon=5\%$\footnote{This value can be tuned lower in real-world applications, resulting in safer but more conservative (and potentially infeasible) policies for all planners.}. The more conservative single-agent planner has a $0\%$ collision rate.

\textbf{Computational time:} We compare the average computation time (see column \textbf{``Comp. T''} in Table~\ref{tab:performance}). The computations were performed on a laptop with an Intel i5-1155G7 CPU. The optimization problems are solved with GUROBI 11.0.1. The \revised{central and single-agent MPC} compute faster, but they require replanning online with a step interval of 0.2s. The LQG-game planner has a longer computational time; However, it computes a state-feedback policy \eqref{eq:feedbackU} offline. The time required for online integration of the policy $u^i_t = \gamma^i_t(x_t)$ is negligible.

\vspace{-1mm}
\begin{table}[hb]
\caption{Average performance over 100 simulations.} \vspace{-1mm}
\centering
\begin{tabular}{| l | r | r | r | r | r |}
    \hline
    \textbf{Method} & \textbf{Cost}  & \textbf{Travel. T} & \textbf{Comp. T}\footnotemark & \textbf{Col. rate} \\ \hline
    LQG-game (SF) & 105.4 & 4.8 s & 3.61 s & 2\%\\ \hline
    \revised{Central-MPC (RH)} & 101.6 & 4.8 s & 0.66 s / step & 2\% \\ \hline
    Single-agent (RH) & 125.4 & 9.6 s & 1.33 s / step & 0\%\\ \hline
\end{tabular} \vspace{-8mm}
\label{tab:performance} 
\end{table}
\footnotetext{For receding-horizon (RH) implementation, \textbf{Comp. T} refers to the online re-planning time at each step (0.2 s). For state-feedback (SF) policies that do not require online re-planning, \textbf{Comp. T} refers to offline computation time.} 

\subsection{Real-world robot experiment}
To examine the real-world performance, we test our algorithm on a robot testbed, consisting of multiple \textit{Nvidia JetBots} running on the Robot Operating System (ROS2). We use an \textit{OptiTrack} motion capture system to measure the states of the robots. We conducted the experiments in a $7 \; \text{m} \times 3.5 \; \text{m}$ area.

Each robot aims to reach its own destination and avoid collision with the others. We linearized the unicycle model of the robots \cite{Keil2020} to formulate the chance-constrained LQG game \eqref{problem:LQGprimal}. With known initial states $x_0$, we use Algorithm \ref{alg:constrainedLQGG} to compute the state-feedback policies $\gamma_{\bm{\bar{\lambda}}}$ offline. At each time step $t$, we use the motion caption camera to observe the state $x_t$ of the multi-robot system and compute the control inputs $u_t^i = \gamma_{\bm{\bar{\lambda}}}^{i}(x_t)$ for all agents. We then convert the control inputs to the wheel rotational speed to actuate the robots in ROS2. The linear state-feedback policy \eqref{eq:feedbackU} enables real-time online computation of the control inputs for the multi-robot system. 
\begin{figure}[!t]
\centering
    \begin{subfigure}
        \centering
        \includegraphics[width=0.493\linewidth]{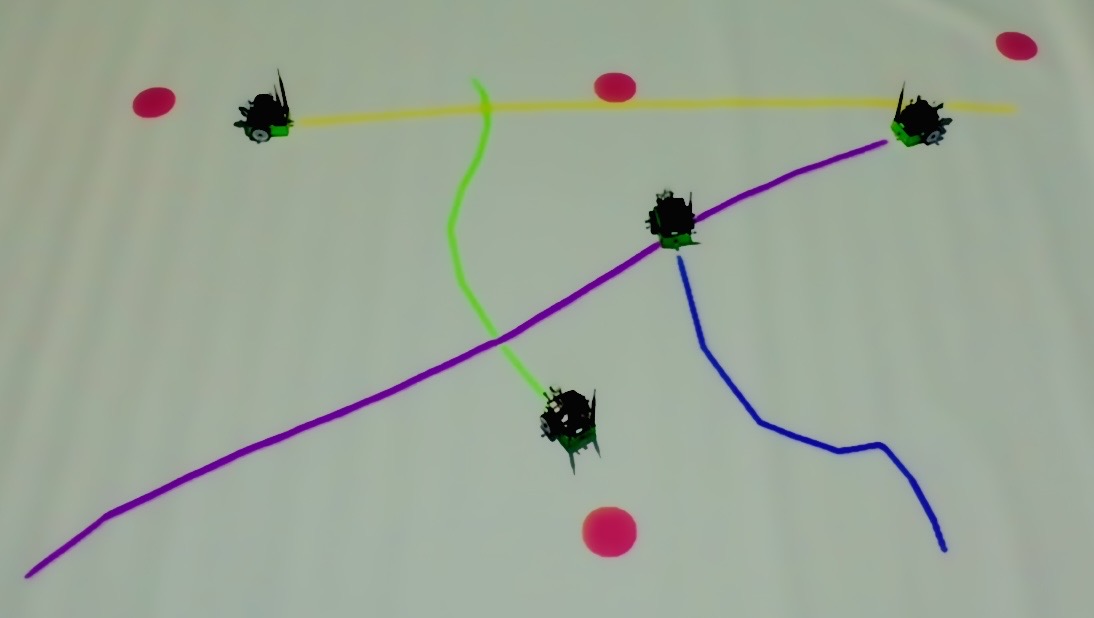}
    \end{subfigure} \hspace{-3mm}
    \begin{subfigure}
        \centering
        \includegraphics[width=0.495\linewidth]{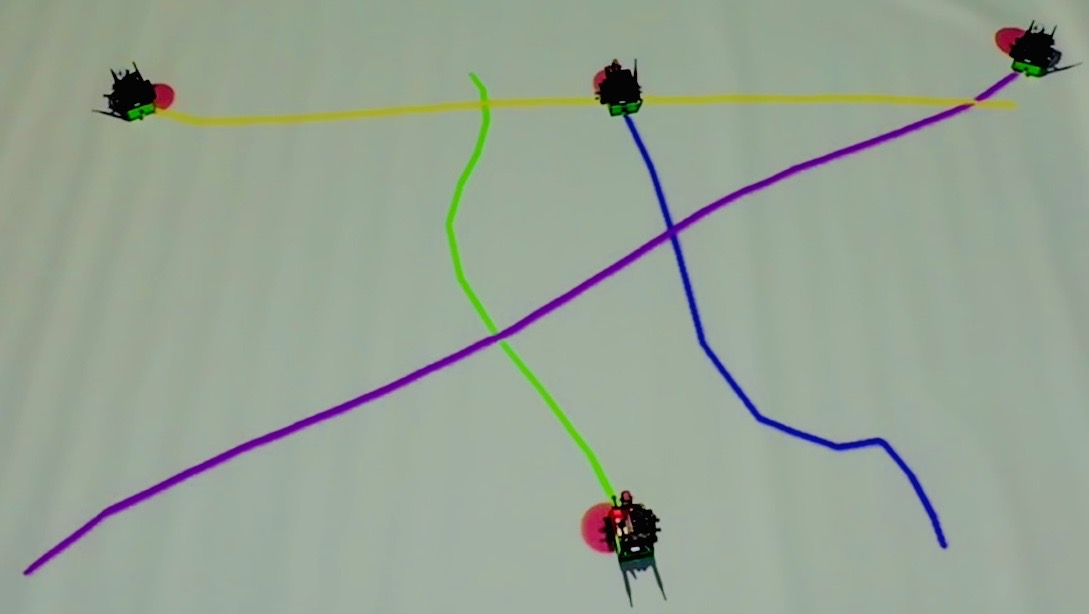}
    \end{subfigure} 
    \vspace{-5mm}
    \caption{Snapshots of a four-robot interaction scenario. The red dots are the goals and the colored lines are the actual paths taken by the robots. With our algorithm, the robots successfully reach their goals and avoid collisions. \href{https://mediaspace.epfl.ch/media/Chance-constrained+linear+quadratic+game+for+multi-robot+interactions/0_fabooxez}{[Link to the videos]}.} 
    \label{fig:four_robots} \vspace{-1mm}
\end{figure} 
We conducted the experiments 50 times with randomly generated goals. The robots constantly get close to their goals with no collisions. A highly interactive scenario with four robots is shown in Fig.~\ref{fig:four_robots}. \vspace{-1mm}

\subsection{Limitations}
Our chance-constrained LQG game planner assumes a linear system with Gaussian uncertainty, which relies on a reference trajectory or iterative local approximation \cite{Keil2020} for linearization. The safety guarantee may fail under non-Gaussian uncertainties, such as those arising from the model mismatch. Additionally, our method assumes full knowledge of other agents' costs and dynamics, making it a central planner that is not robust to incomplete information or communication failures. Future work will focus on relaxing the linear-Gaussian assumption and developing distributed control methods under local and partial information. \vspace{-1mm}

\section{Conclusion} 
We formulated a chance-constrained LQG game to address multi-agent interaction with coupling constraints and system uncertainties. We tractably reformulate the game, in which the chance constraints were approximated conservatively. Thus, any feedback GNE of the reformulated game would ensure the satisfaction of the original constraints. We proposed a dual ascent algorithm and proved its convergence to a feedback GNE of the reformulated game. The algorithm was tested on driving simulations and robot experiments. Our method showed efficacy in ensuring safety under uncertainty and reducing conservativeness for multi-agent systems compared to single-agent trajectory planning. Our future research will focus on relaxing the linear-Gaussian assumption and developing distributed control methods under incomplete information.

\bibliographystyle{IEEEtran}
\bibliography{main}

\begin{thebibliography}{10}
\providecommand{\url}[1]{#1}
\csname url@samestyle\endcsname
\providecommand{\newblock}{\relax}
\providecommand{\bibinfo}[2]{#2}
\providecommand{\BIBentrySTDinterwordspacing}{\spaceskip=0pt\relax}
\providecommand{\BIBentryALTinterwordstretchfactor}{4}
\providecommand{\BIBentryALTinterwordspacing}{\spaceskip=\fontdimen2\font plus
\BIBentryALTinterwordstretchfactor\fontdimen3\font minus \fontdimen4\font\relax}
\providecommand{\BIBforeignlanguage}[2]{{%
\expandafter\ifx\csname l@#1\endcsname\relax
\typeout{** WARNING: IEEEtran.bst: No hyphenation pattern has been}%
\typeout{** loaded for the language `#1'. Using the pattern for}%
\typeout{** the default language instead.}%
\else
\language=\csname l@#1\endcsname
\fi
#2}}
\providecommand{\BIBdecl}{\relax}
\BIBdecl

\bibitem{Basar1998}
T.~Başar and G.~J. Olsder, \emph{{Dynamic Noncooperative Game Theory, 2nd Edition}}.\hskip 1em plus 0.5em minus 0.4em\relax Society for Industrial and Applied Mathematics, 1998.

\bibitem{Facchinei2007}
F.~Facchinei and C.~Kanzow, ``{Generalized Nash Equilibrium Problems},'' \emph{4OR}, vol.~5, no.~3, p. 173–210, 2007.

\bibitem{cleach2021algames}
S.~Le~Cleac’h, M.~Schwager, and Z.~Manchester, ``{ALGAMES: A Fast Augmented Lagrangian Solver for Constrained Dynamic Games},'' \emph{Autonomous Robots}, vol.~46, pp. 201--215, 2022.

\bibitem{Zhu2023}
E.~L. Zhu and F.~Borrelli, ``{A Sequential Quadratic Programming Approach to the Solution of Open-Loop Generalized Nash Equilibria},'' in \emph{2023 IEEE International Conference on Robotics and Automation (ICRA)}, 2023, pp. 3211--3217.

\bibitem{Zhong2023}
H.~Zhong, Y.~Shimizu, and J.~Chen, ``{Chance-Constrained Iterative Linear-Quadratic Stochastic Games},'' \emph{IEEE Robotics and Automation Letters}, vol.~8, no.~1, pp. 440--447, 2023.

\bibitem{Di2020}
B.~Di and A.~Lamperski, ``{Local First-Order Algorithms for Constrained Nonlinear Dynamic Games},'' in \emph{2020 American Control Conference (ACC)}, 2020, pp. 5358--5363.

\bibitem{kavuncu2021potential}
T.~Kavuncu, A.~Yaraneri, and N.~Mehr, ``{Potential iLQR: A Potential-Minimizing Controller for Planning Multi-Agent Interactive Trajectories},'' in \emph{Proceedings of Robotics: Science and Systems}, July 2021.

\bibitem{bhatt2023efficient}
M.~Bhatt, Y.~Jia, and N.~Mehr, ``{Efficient Constrained Multi-Agent Trajectory Optimization Using Dynamic Potential Games},'' in \emph{2023 IEEE/RSJ International Conference on Intelligent Robots and Systems (IROS)}, 2023, pp. 7303--7310.

\bibitem{Li2023}
J.~Li, C.-Y. Chiu, L.~Peters, F.~Palafox, M.~Karabag, J.~Alonso-Mora, S.~Sojoudi, C.~Tomlin, and D.~Fridovich-Keil, ``{Scenario-Game ADMM: A Parallelized Scenario-Based Solver for Stochastic Noncooperative Games},'' in \emph{2023 62nd IEEE Conference on Decision and Control (CDC)}, 2023, pp. 8093--8099.

\bibitem{yadollahi2025}
S.~S. Yadollahi, H.~Kebriaei, and S.~Soudjani, ``{Stochastic Generalized Dynamic Games with Coupled Chance Constraints},'' 2025.

\bibitem{Reddy2017}
P.~V. Reddy and G.~Zaccour, ``{Feedback Nash Equilibria in Linear-Quadratic Difference Games With Constraints},'' \emph{IEEE Transactions on Automatic Control}, vol.~62, no.~2, pp. 590--604, 2017.

\bibitem{Laine2023}
F.~Laine, D.~Fridovich-Keil, C.-Y. Chiu, and C.~Tomlin, ``{The Computation of Approximate Generalized Feedback Nash Equilibria},'' \emph{SIAM Journal on Optimization}, vol.~33, no.~1, pp. 294--318, 2023.

\bibitem{Tsiamis2020CDC}
A.~Tsiamis, D.~S. Kalogerias, L.~F.~O. Chamon, A.~Ribeiro, and G.~J. Pappas, ``{Risk-Constrained Linear-Quadratic Regulators},'' in \emph{2020 59th IEEE Conference on Decision and Control (CDC)}, 2020, pp. 3040--3047.

\bibitem{Zhao2021}
F.~Zhao, K.~You, and T.~Başar, ``{Infinite-horizon Risk-constrained Linear Quadratic Regulator with Average Cost},'' in \emph{2021 60th IEEE Conference on Decision and Control (CDC)}, 2021, pp. 390--395.

\bibitem{Ono2008RA}
M.~Ono and B.~C. Williams, ``{An Efficient Motion Planning Algorithm for Stochastic Dynamic Systems with Constraints on Probability of Failure},'' in \emph{Proceedings of the 23rd Conference on Artificial Intelligence, {AAAI}, Chicago, Illinois, USA, July 13-17, 2008}, 2008, pp. 1376--1382.

\bibitem{Nair2024}
S.~H. Nair, H.~Lee, E.~Joa, Y.~Wang, H.~E. Tseng, and F.~Borrelli, ``{Predictive Control for Autonomous Driving With Uncertain, Multimodal Predictions},'' \emph{IEEE Transactions on Control Systems Technology}, pp. 1--15, 2024.

\bibitem{Altman2000}
E.~Altman and A.~Shwartz, ``{Constrained Markov Games: Nash Equilibria},'' in \emph{Advances in Dynamic Games and Applications}.\hskip 1em plus 0.5em minus 0.4em\relax Boston, MA: Birkh{\"a}user Boston, 2000, pp. 213--221.

\bibitem{ruszczynski2006nonlinear}
A.~Ruszczyński, \emph{Nonlinear Optimization}.\hskip 1em plus 0.5em minus 0.4em\relax Princeton, NJ: Princeton University Press, 2006.

\bibitem{bazaraa2006nonlinear}
M.~S. Bazaraa, H.~D. Sherali, and C.~M. Shetty, \emph{Nonlinear Programming: Theory and Algorithms}, 3rd~ed.\hskip 1em plus 0.5em minus 0.4em\relax John Wiley \& Sons, 2006.

\bibitem{ren2024safe}
K.~Ren, C.~Chen, H.~Sung, H.~Ahn, I.~M. Mitchell, and M.~Kamgarpour, ``{Recursively Feasible Chance-Constrained Model Predictive Control Under Gaussian Mixture Model Uncertainty},'' \emph{IEEE Transactions on Control Systems Technology}, pp. 1--14, 2024.

\bibitem{Keil2020}
D.~Fridovich-Keil, E.~Ratner, L.~Peters, A.~D. Dragan, and C.~J. Tomlin, ``{Efficient Iterative Linear-Quadratic Approximations for Nonlinear Multi-Player General-Sum Differential Games},'' in \emph{IEEE International Conference on Robotics and Automation (ICRA)}, 2020, pp. 1475--1481.

\end{thebibliography}

\appendix  \vspace{-1mm}
\subsection{Proof of Lemma~\ref{lemma:affineInner}} \label{appendix:chanceLinear}
\begin{proof}
Let us define $d_t^{i,j} = x^{i}_{t} - x^{j}_{t} \sim \mathcal{N}(\mu_t^{ij}:= \mu^{i}_{t} - \mu^{j}_{t}, \Sigma_t^{ij}:=\Sigma^{i}_{t} + \Sigma^{j}_{t} + 2\Sigma^{i,j}_{t})$. The covariance of each agent $\Sigma^{i}_{t}$ and the correlation $\Sigma^{i,j}_{t}$ between agents can be obtained from the distribution of $x_t \sim \mathcal{N}(\mu_t, \Sigma_t)$. With a known system dynamics \eqref{eq:systemdynamic}, the state $x_t$'s covariance can be obtained by $
\Sigma_{t+1} = A_t\Sigma_t A_t^\top  + W_t,  \;\;\; \forall t \in [T]^-.$

The collision avoidance constraint can be written as $(d_t^{i,j})^\top C (d_t^{i,j}) \geq R^2.$  A \revised{conservative} affine approximation of the constraint can be found given a reference distance vector $\bar{d}_t^{i,j} = [\bar{r}^{i,j}_{t,1}, \ldots, \bar{r}^{i,j}_{t,n_x}]^\top$, where $||\bar{d}_t^{i,j}||_C = R$. As $d_t^{i,j}$ is Gaussian distributed, the chance constraints \eqref{constraint:collisionavoidance} can be deterministically reformulated as follows \revised{\cite{Ono2008RA, Nair2024}}: \vspace{-3mm}

{\small
\begin{subequations}
\begin{alignat}{2}
    \eqref{constraint:collisionavoidance} & \; \Leftrightarrow \; \mathbb{P}\left((d_t^{i,j})^\top C (d_t^{i,j}) \geq R^2\right) \geq 1 - \epsilon_t^k \\
     \Leftarrow & \;\; \mathbb{P}\left(2C\bar{d}_t^{i,j^\top}(d_t^{i,j} - \bar{d}_t^{i,j}) \geq 0\right) \geq 1 - \epsilon_t^k \label{subeq:affineInner}\\
    \Leftrightarrow & \;\; C\bar{d}_t^{i,j^\top}[\mu_t^{ij} - \bar{d}_t^{i,j}] + \text{\textPsi}^{-1}(1 - \epsilon_t^k) \|C\bar{d}_t^{i,j}\|_{\Sigma_t^{ij}} \leq 0 \\
    \equiv & \;\; l_t^{i,j^\top}  \bbE \left[x_t\right] + c_t^{i,j} \leq 0 \label{constraint:CCaffine},
\end{alignat}
\end{subequations}}where $\text{\textPsi}^{-1}(\cdot)$ denotes the inverse cumulative distribution function of $\mathcal{N}(0, 1)$. Then the decoupled box constraints \eqref{eq:boxStateConstraint} can be reformulated as $l_t^{i,q^\top}  \bbE \left[x_t\right] + c_t^{i,q} \leq 0.$ The reformulation follows similarly to $\eqref{subeq:affineInner} \Leftrightarrow \eqref{constraint:CCaffine}$. Finally, the linear constraints reformulated from \eqref{eq:boxStateConstraint} and \eqref{constraint:collisionavoidance} across all agents collectively result in 
$M$ constraints, which can be compactly expressed as $\vl^\top \vx + \vc \leq 0$ with $\vl \in \bbR^{T n_x \times M}$ and $\vc \in \bbR^M$.
\end{proof} \vspace{-3mm}

\subsection{Parameters for finite-horizon LQG games} \label{appendix:LQfeedback}

Based on \cite[Remark 6.3]{Basar1998}, the feedback parameters of each player $i$ at time $t$ is \vspace{-6mm}

{\footnotesize\begin{equation} \label{eq:stateFeedbackGain}
\begin{aligned}
    K^{i*}_t = \left(R_{t}^{i} + (B_{t}^{i})^\top P^{i*}_{t+1} B_{t}^{i} \right)^{-1} (B_{t}^{i})^\top P^{i*}_{t+1} \left( A_t - \sum_{j \neq i} B_t^j K_t^{j*} \right), \\
    \alpha^{i*}_t = \left(R_{t}^{i} + (B_{t}^{i})^\top P^{i*}_{t+1} B_{t}^{i} \right)^{-1} (B_{t}^{i})^\top \left( \zeta^{i*}_{t+1} - P^{i*}_{t+1} \sum_{j \neq i} B_t^j \alpha_t^{j*} \right),
\end{aligned}
\end{equation}}

\noindent where $P^{i*}_t$ and $\zeta^{i*}_t$ can be obtained recursively backward from the coupled Riccati equations with augmented cost matrices: \vspace{-3mm}

{\small
\begin{align}
    &\textstyle P^{i*}_t = \left( A_t - \sum_{j = 1}^N B_t^j K_t^{j*} \right)^\top P^i_{t+1} \left( A_t - \sum_{j = 1}^N B_t^j K_t^{j*} \right) \notag\\
    & \quad \quad \quad + (K^{i*}_t)^\top R_t^i K^{i*}_t  + Q_{t}^{i}, \notag \\
    &\textstyle \zeta^{i*}_t = \left( A_t - \sum_{j = 1}^N B_t^j K_t^{j*} \right)^\top \left( \zeta^{i*}_{t+1} - P^i_{t+1} \sum_{j = 1}^N B_t^j \alpha_t^{j*} \right) \notag \\
    & \textstyle \quad \quad \quad + \sum_{j = 1}^N (K^{j*}_t)^\top R_t^j \alpha^{j*}_t + l_{t} \bm{\lambda}, \notag \\ \vspace{2mm}
    & \textstyle P^{i*}_T = Q_{T}^{i}, \text{ and }   \zeta^{i*}_T = l_{T} \bm{\lambda}. \label{eq:coupledRE}
\end{align}} \vspace{-5mm}

\noindent Above, $l_t \in \bbR^{n_x \times M}$ denotes the set of constraint parameters corresponding to time $t$. Note that \cite[Remark 6.5]{Basar1998} derived a sufficient condition for the unique solvability of \eqref{eq:stateFeedbackGain}. \vspace{-3mm}

\subsection{Proof of lemma~\ref{lemma:dualGradient}} \label{appendix:proofTheorem}
\begin{proof}
(1) Recall that $\cD^i (\bm{\lambda}) = \cD^i (\bm{\lambda}; \gamma^{-i}_{\bm{\lambda}})$. First, we show that $\nabla\cD^i (\bm{\lambda}) = \vg (\vx_{\bm{\lambda}})$ based on \cite[Theorem 6.3.3]{bazaraa2006nonlinear}.
For the clarity of the proof, we denote the constraint values induced by policy $(\gamma^i, \gamma^{-i})$ as $\vg(\gamma^i, \gamma^{-i})$. At the Nash equilibrium (NE) $(\gamma^i_{\bm{\lambda}}, \gamma^{-i}_{\bm{\lambda}})$ corresponding to an arbitrary $\bm{\lambda}$,  $\vg (\vx_{\bm{\lambda}}) = \vg(\gamma^i_{\bm{\lambda}}, \gamma^{-i}_{\bm{\lambda}})$. For \(\bm{\lambda}, \bm{\lambda'} \in \bbR^M_{\geq 0}\), we define \vspace{-1mm}
{\small
\begin{align} \vspace{-5mm}
    &\gamma^i_{\bm{\lambda}} = \arg\min_{\gamma^i } \cJ(x_0, \gamma^i, \gamma^{-i}_{\bm{\lambda}}) + \bm{\lambda}^\top \vg(\gamma^i, \gamma^{-i}_{\bm{\lambda}}), \label{eq:minLambda}\\
    &\gamma^i_{\bm{\lambda'}} = \arg\min_{\gamma^i } \cJ(x_0, \gamma^i, \gamma^{-i}_{\bm{\lambda}}) + \bm{\lambda'}^\top \vg(\gamma^i, \gamma^{-i}_{\bm{\lambda}}).
\end{align}} \vspace{-4mm}

\revised{From \eqref{eq:minLambda},} it follows that for all $i \in [N]$: \vspace{-4mm}

{\footnotesize
\begin{align}
    & D^i(\bm{\lambda}; \gamma^{-i}_{\bm{\lambda}}) - D^i(\bm{\lambda'}; \gamma^{-i}_{\bm{\lambda}}) \leq \cJ^i(x_0, \gamma^i_{\bm{\lambda'}}, \gamma^{-i}_{\bm{\lambda}}) + \bm{\lambda}^\top \vg(\gamma^i_{\bm{\lambda'}}, \gamma^{-i}_{\bm{\lambda}}) \notag\\ 
    & \quad\quad\quad \quad \quad \quad \quad \quad \quad \quad \quad \quad - \cJ^i(x_0, \gamma^i_{\bm{\lambda'}}, \gamma^{-i}_{\bm{\lambda}}) - \bm{\lambda'}^\top \vg(\gamma^i_{\bm{\lambda'}}, \gamma^{-i}_{\bm{\lambda}}) \notag\\
    & \quad \quad \quad \quad \quad \quad \quad \quad \quad \quad\quad = (\bm{\lambda} - \bm{\lambda'})^\top \vg(\gamma^i_{\bm{\lambda'}}, \gamma^{-i}_{\bm{\lambda}}). \label{eq:dualineq1}\\
    & \text{\normalsize Similarly, } D^i(\bm{\lambda'}; \gamma^{-i}_{\bm{\lambda}}) - D^i(\bm{\lambda}; \gamma^{-i}_{\bm{\lambda}}) \leq (\bm{\lambda'} - \bm{\lambda})^\top \vg(\gamma^i_{\bm{\lambda}}, \gamma^{-i}_{\bm{\lambda}}). \label{eq:dualineq2}
\end{align}} \vspace{-5mm}

By rearranging the terms of \eqref{eq:dualineq2}, we can get \vspace{-3mm}

{\small
\begin{align}
0 & \geq D^i(\bm{\lambda'}; \gamma^{-i}_{\bm{\lambda}}) - D^i(\bm{\lambda}; \gamma^{-i}_{\bm{\lambda}}) - (\bm{\lambda'} - \bm{\lambda})^\top \vg(\gamma^i_{\bm{\lambda}}, \gamma^{-i}_{\bm{\lambda}}) \notag\\
& \geq (\bm{\lambda'} - \bm{\lambda})^\top \left[\vg(\gamma^i_{\bm{\lambda'}}, \gamma^{-i}_{\bm{\lambda}}) - \vg(\gamma^i_{\bm{\lambda}}, \gamma^{-i}_{\bm{\lambda}}) \right] \notag\\
& \geq -\|\bm{\lambda'} - \bm{\lambda}\| \|\vg(\gamma^i_{\bm{\lambda'}}, \gamma^{-i}_{\bm{\lambda}}) - \vg(\gamma^i_{\bm{\lambda}}, \gamma^{-i}_{\bm{\lambda}})\|. \label{eq:absIneq}
\end{align}}

The second inequality holds due to \eqref{eq:dualineq1}. The third inequality holds due to the Cauchy–Schwarz inequality. By rearranging the terms of \eqref{eq:absIneq}, we can get \vspace{-3mm}

{\small
\begin{align*}
0 & \geq \frac{D^i(\bm{\lambda'}; \gamma^{-i}_{\bm{\lambda}}) - D^i(\bm{\lambda}; \gamma^{-i}_{\bm{\lambda}}) - (\bm{\lambda'} - \bm{\lambda})^\top \vg(\gamma^i_{\bm{\lambda}}, \gamma^{-i}_{\bm{\lambda}})}{\|\bm{\lambda'} - \bm{\lambda}\|} \\
& \geq -\|\vg(\gamma^i_{\bm{\lambda'}}, \gamma^{-i}_{\bm{\lambda}}) - \vg(\gamma^i_{\bm{\lambda}}, \gamma^{-i}_{\bm{\lambda}})\|.
\end{align*}}

As \(\bm{\lambda'} \to \bm{\lambda}\), \(\vg(\gamma^i_{\bm{\lambda}}, \gamma^{-i}_{\bm{\lambda}}) \to \vg(\gamma^i_{\bm{\lambda'}}, \gamma^{-i}_{\bm{\lambda}})\). Therefore,
{\small
\begin{equation*}
\lim_{\bm{\lambda'} \to \bm{\lambda}} \frac{D^i(\bm{\lambda'}; \gamma^{-i}_{\bm{\lambda}}) - D^i(\bm{\lambda}; \gamma^{-i}_{\bm{\lambda}}) - (\bm{\lambda'} - \bm{\lambda})^\top \vg(\gamma^i_{\bm{\lambda}}, \gamma^{-i}_{\bm{\lambda}})}{\|\bm{\lambda'} - \bm{\lambda}\|} = 0.
\end{equation*}}

This concludes that \(\cD^i(\bm{\lambda})\) is differentiable and \(\nabla \cD^i (\bm{\lambda}) = \vg(\gamma^i_{\bm{\lambda}}, \gamma^{-i}_{\bm{\lambda}}) = \vg (\vx_{\bm{\lambda}})\). \vspace{1mm}

(2) Next, we show that $\vg (\vx_{\bm{\lambda}}) = \tilde{L}^\top \bm{\lambda} + \vc$. By rearranging and stacking over all agents' $\alpha^{i*}_t$ in \eqref{eq:stateFeedbackGain}, we obtain
{\footnotesize \begin{align*}
    \setlength{\arraycolsep}{1pt}
    \begin{bmatrix} R_{t}^{1} + (B_{t}^{1})^\top P^{1*}_{t+1} B_{t}^{1} & \quad \ldots & \quad(B_{t}^{1})^\top P^{1*}_{t+1} B^N_t  \\
    \vdots & \quad \ddots & \vdots \\
    (B_{t}^{N})^\top P^{N*}_{t+1} B^1_t & \quad \ldots & \quad R_{t}^{N} + (B_{t}^{N})^\top P^{N*}_{t+1} B_{t}^{N}
    \end{bmatrix} \begin{bmatrix}
        \alpha^{1*}_t \\ \vdots \\ \alpha^{N*}_t
    \end{bmatrix} \\ 
    \quad \quad = \begin{bmatrix} (B_{t}^{1})^\top & \ldots & 0\\
        \vdots &  \ddots & \vdots\\  0 & \ldots & (B_{t}^{N})^\top 
    \end{bmatrix} \begin{bmatrix}
        \zeta^{1*}_{t+1} \\ \vdots \\ \zeta^{N*}_{t+1} 
    \end{bmatrix}.
\end{align*}} 

Stacking $\alpha^*_t$ and $\zeta^*_t$ over all $t \in [T]$, we get a compact expression $\alpha^* = H \zeta^*$. Let us define $F_t = A_t - \sum_{i=1}^N B_t^i K_t^{i*} \in \mathbb{R}^{n_x\times n_x}$. Recall that $\zeta^*$ can be solved backward via~\eqref{eq:coupledRE}. For an arbitrary $\bm{\lambda}$, by stacking $\zeta^{i*}_t$ over all agents, we get \vspace{-2mm}

{\scriptsize \setlength{\arraycolsep}{3.6pt}
\begin{align*}
    & \begin{bmatrix}
        \zeta^{1*}_{t} \\ \vdots \\ \zeta^{N*}_{t} 
    \end{bmatrix}  = \begin{bmatrix} (F_{t})^\top & \ldots & 0 \\
                        \vdots & \ddots & \vdots\\ 0 & \ldots & (F_{t})^\top 
                    \end{bmatrix}  \begin{bmatrix}
                        \zeta^{1*}_{t+1} \\ \vdots \\ \zeta^{N*}_{t+1} \end{bmatrix} + \begin{bmatrix} l_{t}  \\
                        \vdots \\ l_{t} 
                    \end{bmatrix} \bm{\lambda} \\
                     & \quad + \begin{bmatrix} 
    K_{t}^{1*}R_t^1-F_t^\top P_{t+1}^{1*} B_t^1 & \ldots & K_{t}^{N*}R_t^N-F_t^\top P_{t+1}^{1*} B_t^N  \\
    \vdots & \ddots & \vdots \\
    K_{t}^{1*}R_t^1-F_t^\top P_{t+1}^{N*} B_t^1 & \ldots & K_{t}^{N*}R_t^N-F_t^\top P_{t+1}^{N*} B_t^N \end{bmatrix} \begin{bmatrix}
                    \alpha^{1*}_{t} \\ \vdots \\ \alpha^{N*}_{t}  \end{bmatrix}.
\end{align*}}

Since $\zeta^{i*}_T = l_{T} \bm{\lambda}$, we get $\zeta^*$ is a linear function of $\bm{\lambda}$. Hence, $\alpha^* = G \bm{\lambda}$ for some $G \in \bbR^{TNn_u \times M}$. Let us denote $B_t =  [B_t^1, \cdots, B_t^N]$. 
By integrating the expected states, we get $\bbE(x_1) = F_0 x_0 - B_0\alpha_0^*, \; \bbE(x_2) = F_1 \bbE(x_1) - B_1 \alpha_1^*, ...$ and so on. One can verify that, for an arbitrary $\bm{\lambda}$, the expected state trajectory can be written as $\vx_{\bm{\lambda}} = [I, \; F_0, \; F_1F_0, \; ... \;]^\top x_0 + ZG\bm{\lambda}$ for some $Z \in \mathbb{R}^{ (T+1) n_x \times TN n_u}$. Hence, the constraints \eqref{constraint:primalAffine} can be written as $\vg(\vx_{\bm{\lambda}}) =\vl^{\top} \vx_{\bm{\lambda}} + \vc \equiv \tilde{L}^\top \bm{\lambda} + \tilde{\vc}.$
\end{proof}

\end{document}